\documentclass{article}

\usepackage[final]{neurips_2019}

\usepackage[utf8]{inputenc} % allow utf-8 input
\usepackage[T1]{fontenc}    % use 8-bit T1 fonts
\usepackage{hyperref}       % hyperlinks
\usepackage{url}            % simple URL typesetting
\usepackage{booktabs}       % professional-quality tables
\usepackage{amsfonts}       % blackboard math symbols
\usepackage{nicefrac}       % compact symbols for 1/2, etc.
\usepackage{microtype}      % microtypography
\usepackage{mathtools}

\usepackage{amsfonts}
\usepackage{amsthm}
\usepackage{amsmath}
\usepackage{amssymb}

\usepackage{algorithm,algpseudocode}
\usepackage{wrapfig}
\usepackage[colorinlistoftodos]{todonotes}

\newtheorem{proposition}{Proposition}

\newtheorem{example}{Example}

\theoremstyle{definition}
\newtheorem{definition}{Definition}

\newtheorem{remark}{Remark}

\newcommand{\lie}{\mathsf{L}_{f_u}}
\begin{document}
%\todo[inline]{Draft. Do not distribute.}

\title{Neural Lyapunov Control}

% The \author macro works with any number of authors. There are two
% commands used to separate the names and addresses of multiple
% authors: \And and \AND.
%
% Using \And between authors leaves it to LaTeX to determine where to
% break the lines. Using \AND forces a line break at that point. So,
% if LaTeX puts 3 of 4 authors names on the first line, and the last
% on the second line, try using \AND instead of \And before the third
% author name.

\author{
  Ya-Chien Chang\\UCSD\\yac021@eng.ucsd.edu
  \And Nima Roohi\\UCSD\\nroohi@eng.ucsd.edu \And Sicun Gao\\
  UCSD\\sicung@eng.ucsd.edu
  %% examples of more authors
  %% \And
  %% Coauthor \\
  %% Affiliation \\
  %% Address \\
  %% \texttt{email} \\
  %% \AND
  %% Coauthor \\
  %% Affiliation \\
  %% Address \\
  %% \texttt{email} \\
  %% \And
  %% Coauthor \\
  %% Affiliation \\
  %% Address \\
  %% \texttt{email} \\
  %% \And
  %% Coauthor \\
  %% Affiliation \\
  %% Address \\
  %% \texttt{email} \\
}

\maketitle

\begin{abstract}
We propose new methods for learning control policies and neural network Lyapunov functions for nonlinear control problems, with provable guarantee of stability. The framework consists of a learner that attempts to find the control and Lyapunov functions, and a falsifier that finds counterexamples to quickly guide the learner towards solutions. The procedure terminates when no counterexample is found by the falsifier, in which case the controlled nonlinear system is provably stable. The approach significantly simplifies the process of Lyapunov control design, provides end-to-end correctness guarantee, and can obtain much larger regions of attraction than existing methods such as LQR and SOS/SDP. We show experiments on how the new methods obtain high-quality solutions for challenging control problems. %such as path tracking for wheeled vehicles and humanoid robot balancing. 
\end{abstract}

\section{Introduction}

Learning-based methods hold the promise of solving hard nonlinear control problems in robotics. Most existing work focuses on learning control functions represented as neural networks through repeated interactions of an unknown environment in the framework of deep reinforcement learning, with notable success. However, there are still well-known issues that impede the immediate use of these methods in practical control applications, including sample complexity, interpretability, and safety~\cite{DBLP:journals/corr/AmodeiOSCSM16}. Our work investigates a different direction: Can learning methods be valuable even in the most classical setting of nonlinear control design? We focus on the challenging problem of designing feedback controllers for stabilizing nonlinear dynamical systems with provable guarantee. This problem captures the core difficulty of underactuated robotics~\cite{russbook}. We demonstrate that neural networks and deep learning can find provably stable controllers in a direct way and tackle the full nonlinearity of the systems, and significantly outperform existing methods based on linear or polynomial approximations such as linear-quadratic regulators (LQR)~\cite{Kwakernaak} and sum-of-squares (SOS) and semidefinite programming (SDP)~\cite{Parrilo2000}. The results show the promise of neural networks and deep learning in improving the solutions of many challenging problems in nonlinear control. 

%even without enlarging the control policy space and fully known dynamics? We show in this paper that even for problems where the system dynamics is known and the control function is simple, learning based methods can achieve impressive results and provide new solutions to core nonlinear control problems and outperform existing methods. Our approach relies on the Lyapunov stability theory for nonlinear control, for which we deep learn a Lyapunov function as well as the corresponding control policy, so that the full controlled system is fully guaranteed to be stable with a sizable region of attraction. This work is the first in showing that neural Lyapunov function can provide full guarantee of stability and good region of attraction. 
%A wide range of challenging control design problems are about restricted control policies with explicitly given dynamics models. Existing methods can only work under conservative assumptions that bring the system close to linear systems so that the system behaviors are analyzable, accomplishing only a fraction of the tasks that can be achieved by the motor capacities~\cite{underacuated}. 

%As is well-known, nonlinear control is largely an open problem with the large control and robotics community working on designing simple reactive control functions for nonlinear systems. without enriching the control function class, and provide a new framework for solving hard nonlinear control problems with full reliability? 

The prevalent way of stabilizing nonlinear dynamical systems is to linearize the system dynamics around an equilibrium, and formulate LQR problems to minimize deviation from the equilibrium. LQR methods compute a linear feedback control policy, with stability guarantee within a small neighborhood where linear approximation is accurate. However, the dependence on linearization produces extremely conservative systems, and it explains why agile robot locomotion is hard~\cite{russbook}. To control nonlinear systems outside their linearizable regions, we need to rely on Lyapunov methods~\cite{lyapunovbook}. Following the intuition that a dynamical system stabilizes when its energy decreases over time, Lyapunov methods construct a scalar field that can force stabilization. These fields are highly nonlinear and the need for function approximations has long been recognized~\cite{lyapunovbook}. Many existing approaches rely on polynomial approximations of the dynamics and the search of sum-of-squares polynomials as Lyapunov functions through semidefinite programming (SDP)~\cite{Parrilo2000}. A rich theory has been developed around the approach, but in practice the polynomial approximations pose much restriction on the systems and the Lyapunov landscape. Moreover, well-known numerical sensitivity issues in SDP~\cite{Lofberg2009} make it very hard to find solutions that fully satisfy the Lyapunov conditions. In contrast, we exploit the expressive power of neural networks, the convenience of gradient descent for learning, and the completeness of nonlinear constraint solving methods to provide full guarantee of Lyapunov conditions. We show that the combination of these techniques produces control designs that can stabilize various nonlinear systems with verified regions of attraction that are much larger than what can be obtained by existing control methods. 

%Indeed, as we will shown in the experiments section, even when SOS Lyapunov functions are found by the standard algorithms, simulation shows violation of 
%have always been core in Lyapunov methods for nonlinear control. The standard approach is to use nonlinear polynomials (typically sum-of-squares) to construct Lyapunov functions. The functions are computed through symbolic and numerical search.~\todo{talk about how local LQR is} Typically polynomials are used for capturing the nonlinearity. Polynomials have nice symbolic properties when quadratic candiates work, but their properties quickly become unmanagable as degree increases, and their parameters grow exponentially. Recent work has started using learning to construct Lyapunov functions using neural networks. 
%It has been shown that specially constructed neural networks can be used to approximate the region of attraction based on sampling the state space with a fixed control law. These Lyapunov functions are not certified to satisfy the standard Lyapunov conditions, and can only provide approximations to a discretized version of the original problem, and assume that the control functions are already given. 

We propose an algorithmic framework for learning control functions and neural network Lyapunov functions for nonlinear systems without any local approximation of their dynamics. The framework consists of a learner and a falsifier. The learner uses stochastic gradient descent to find parameters in both a control function and a neural Lyapunov function, by iteratively minimizing the {\em Lyapunov risk} which measures the violation of the Lyapunov conditions. The falsifier takes a control function and Lyapunov function from the learner, and searches for {\em counterexample} state vectors that violate the Lyapunov conditions. The counterexamples are added to the training set for the next iteration of learning, generating an effective curriculum. The falsifier uses delta-complete constraint solving~\cite{DBLP:conf/cade/GaoAC12}, which guarantees that when no violation is found, the Lyapunov conditions are guaranteed to hold for all states in the verified domain. In this framework, the learner and falsifier are given tasks that are difficult in different ways and can not be achieved by the other side. Moreover, we show that the framework provides the flexibility for fine-tuning the control performance by directly enlarging the region of attraction on demand, by adding regulator terms in the learning cost. 

We experimented with several challenging nonlinear control problems in robotics, such as drone landing, wheeled vehicle path following, and n-link planar robot balancing\cite{nlinkrobot}. We are able to find new control policies that produce certified region of attractions that are significantly larger than what can be established previously. We provide a detailed analysis of the performance comparison between the proposed methods and the LQR/SOS/SDP methods. 
%learning control functions and neural network Lyapunov function. A falsifier for  A tunable component for iteratively enlarging the region of attraction. ...
%interaction edemonstrate that it is possible to learn both control functions neural network Lyapunov functions fully automatically on nonlinear continuous problems, and guarantee that the Lyapunov conditions hold for all states. The learned Lyapunov functions can provide region of attractions that are siginificantly larger than what the existing control methods can find, and thus finding controllers that give much better (stable and safe) performance. Our framework consists of the following components: 
%We emphasize that the methods are numerically robust. We demonstrate the effectiveness of the methods in various hard control problems. We first use the well-known inverted pendulum pendulum to compare the performance of controllers designed through... we show that the region of attraction obtained for simple controllers is large...
%The paper is organized as follows. 

\noindent{\bf Related Work.} 
\newcommand{\etal}{{\em et.\ al.}}%
%Compared to the control-theoretic approaches, neural Lyapunov control provides a much simpler design process, relying purely on gradient-based methods for the learning. The saving is similar to the reduction of feature engineering and specific optimization methods in other areas of AI. 
The recent work of Richards~\etal~\cite{Spencer2018} has also proposed and shown the effectiveness of using neural networks to learn safety certificates in a Lyapunov framework, but our goals and approaches are different. Richards~\etal\ focus on discrete-time nonlinear systems and the use of neural networks to learn the region of attraction of a given controller. The Lyapunov conditions are validated in relaxed forms through sampling. Special design of the neural architecture is required to compensate the lack of complete checking over all states. In comparison, we focus on learning the control and the Lyapunov function together with provable guarantee of stability in larger regions of attraction. Our approach directly handles continuous nonlinear dynamical systems, does not assume control functions are given other than an initialization, and uses generic feed-forward network representations without manual design. Our approach successfully works on many more nonlinear systems, and find new control functions that enlarge regions of attraction obtainable from standard control methods. Related learning-based approaches for finding Lyapunov functions include%
%~\cite{Vannelli1985,Venkat1992,khalil2002,Silva2005}
~\cite{Berkenkamp2016,Berkenkamp2017,Chow2018,Rasmussen2006}. There is strong evidence that linear control functions are all we need for solving highly nonlinear control problems through reinforcement learning as well~\cite{simple-linear}, suggesting convergence of different learning approaches. In the control and robotics community, similar learner-falsifier frameworks have been proposed by~\cite{Ravanbakhsh2019,Kapinski2014} without using neural network representations. The common assumption is the Lyapunov functions are high-degree polynomials. In these methods, an explicit control function and Lyapunov function can not be learned together because of the bilinear optimization problems that they generate. Our approach significantly simplifies the algorithms in this direction and has worked reliably on much harder control problems compared to existing methods. Several theoretical results on asymptotic Lyapunov stability~\cite{Ahmadi2011,Ahmadi2012,Ahmadi2013,Ahmadi2015} show that some very simple dynamical systems do not admit a polynomial Lyapunov function of any degree, despite being globally asymptotically stable. Such results further motivates the use of neural networks as a more suitable function approximator. A large body of work in control uses SOS representations and SDP optimization in the search for Lyapunov functions~\cite{Henrion2005,Parrilo2000,Chesi2009,Jarvis2003,Majumdar2017}. However, scalability and numerical sensitivity issues have been the main challenge in practice. As is well known, the number of semidefinite programs from SOS decomposition grows quickly for low degree polynomials~\cite{Parrilo2000}. 

\section{Preliminaries}
%We make extensive use of the following results from nonlinear control theory and Lyapunov stability theory. 
We consider the problem of designing control functions to stablize a dynamical system at an equilibrium point. We make extensive use of the following results from Lyapunov stability theory. 
\begin{definition}[Controlled Dynamical Systems]
\label{def:dyn}
An $n$-dimensional controlled dynamical system is 
\begin{equation}
\label{eqn:dynamics}
    \frac{\mathrm{d}x}{\mathrm{d}t}=f_{u}(x), \quad x(0)=x_{0} 
\end{equation}
where $f_u:\mathcal{D}\rightarrow \mathbb{R}^{n}$ is a Lipschitz-continuous vector field, and $\mathcal{D}\subseteq \mathbb{R}^{n}$ is an open set with $0\in \mathcal{D}$ that defines the state space of the system. Each $x(t) \in \mathcal{D}$ is a state vector. The feedback control is defined by a continuous function $u:\mathbb{R}^{n}\rightarrow \mathbb{R}^{m}$, used as a component in the full dynamics $f_u$.
\end{definition}%, $u \in R^{m}$ is the control inputs vector. 

%Throughout the paper, we study stabilization of nonlinear dynamical systems within a domain of interest using control Lyapunov functions (CLFs). 
%\todo[inline]{we'll use Lyapunov functions, not control Lyapunov functions}

\begin{definition}[Asymptotic Stability]
We say that system of $(1)$ is stable at the origin if for any $\varepsilon\in\mathbb{R}^+$, there exists $\delta(\varepsilon) \in\mathbb{R}^+$ such that if $\|x(0)\|<\delta$ then $\|x(t)\|<\varepsilon$ for all $t\geq 0$. The system is asymptotically stable at the origin if it is stable and also $\lim_{t\rightarrow\infty}\|x(t)\|=0$ for all $\|x(0)\|<\delta$. \end{definition}
%The control design problem is about finding a control function $u$ such that the system is stable around an equilibrium point. Without loss of generality, we assume the system has an equilibrium point at origin $0$ that we seek to stabilize to. 
\begin{definition}[Lie Derivatives]\label{def:lie} 
The Lie derivative of a continuously differentiable scalar function $V:\mathcal{D}\rightarrow \mathbb{R}$ over a vector field $f_u$ is defined as \[\lie V\left({x}\right)=\sum_{i=1}^{n} \frac{\partial V}{\partial x_{i}}\frac{\mathrm{d} x_i}{\mathrm{d} t}= \sum_{i=1}^{n} \frac{\partial V}{\partial x_{i}}[f_{u}]_i(x)\] 
It measures the rate of change of $V$ along the direction of the system dynamics.
\end{definition}
\begin{proposition}[Lyapunov Functions for Asymptotic Stability]\label{def:lya}
Consider a controlled system $(1)$ with equilibrium at the origin, i.e., $f_u(0)=0$. Suppose there exists a continuously differentiable function $V:\mathcal{D}\rightarrow \mathbb{R}$ that satisfies the following conditions: 
\begin{equation}
    V\left(0\right)= 0 \mbox{, and, } \forall x \in \mathcal{D}\setminus\{0\},  V\left({x}\right)> 0\mbox{ and }\lie V\left({x}\right)<0.
\end{equation}
Then, the system is asymptotically stable at the origin and $V$ is called a Lyapunov function. 
\end{proposition}
%The last condition guarantees $V\left(x\right)$ decreases along the trajectories of the given system within $\mathcal{D}$, i.e., by applying a proper feedback control $u$ any initial state within $\mathcal{D}$ must eventually get to the origin.
Linear-Quadratic Regulators (LQR) is a widely-adpoted optimal control strategy. LQR controllers are guaranteed to work within a small neighborhood around the stationary point where the dynamics can be approximated as linear systems. A detailed description can be found in~\cite{Kwakernaak}.  
%However, ensuring large region of attraction in the controlled systems requires us to find Lyapunov functions to show that the linear control can also work with the nonlinear dynamics. This is the core difficulty of nonlinear control. 
%\begin{definition}[LQR]
%Consider a dynamics $\dot{x}=Ax+Bu$ obtained by linearizing the system $(1)$ around a stationary point is controllable and define a cost function:  
%\[J =\int_{0}^{\infty}\left[x^{T}(t)Q x(t)+u^{T}(t)R u(t)\right] d t\] 
%, where $Q$ and $R$ are symmetric, positive semidefinite matrices. Given the optimal feedback controller that minimizes the value of the cost takes the form: $u^{*}=-R^{-1}B^{T}Sx = -Kx$, then $S$ can be found by solving the algebraic Riccati equation: $Q-SBR^{-1}B^{T}S+SA+A^{T}S=0$. $u^{x}=-Kx$ defines the optimal LQR controller.
%\end{definition}

\section{Learning to Stabilize with Neural Lyapunov Functions}

We now describe how to learn both a control function and a neural Lyapunov function together, so that the Lyapunov conditions can be rigorously verified to ensure stability of the system. 
%This section describes our approach to synthesize a neural control Lyapunov function for a given dynamic system. 
%The algorithm consists of two main components: a learner that formulates the control policy and the corresponding neural network Lyapunov function, and a falsifier that attempts to find counterexamples in the state space that witness the violation of the Lyapunov conditions. When counterexamples are found, they are augmented with more samples from their neighborhoods and sent back to the learner for the next iteration. The process terminates when the falsifier can not find any counterexamples, in which case the learned control policy and Lyapunov function are guaranteed to ensure stability. This guarantee is made possible by the use of SMT solving algorithms that ensure delta-completeness~\cite{DBLP:conf/cade/GaoAC12}. 
We provide pseudocode of the algorithm in Algorithm~\ref{alg:clf}.
%learning and verification, illustrated in Figure $1$. A candidate neural CLF is obtained by training neural network and the candidate is verified by using dReal solver If dReal returns a counterexample that violates CLF conditions, the counterexample will be added to the training dataset for improving learning for generating a neural CLF. Otherwise, the candidate is proved to be a valid CLF.

\subsection{Control and Lyapunov Function Learning}
% Design of the cost function. 
% LQR part.

We design the hypothesis class of candidate Lyapunov functions to be multilayered feedforward networks with $\tanh$ activation functions. It is important to note that unlike most other deep learning applications, we can not use non-smooth networks, such as with ReLU activations. This is because we will need to analytically determine whether the Lyapunov conditions hold for these neural networks, which requires the existence of their Lie derivatives. 
%properties of these networks will be analyzed we aim for rigorous analysis if the Lie derivatives of the Lyapunov functions, ...

%representation of the candidate Lyapunnov function, as a feed-forward 
%of a given dynamical system is a feed-forward neural network with $M-1$ hidden layers as Figure 1. 
For a neural network Lyapunov function, its input is any state vector of the system in Definition (\ref{def:dyn}) and the output is a scalar value. We write $\theta$ to denote the parameter vector for a Lyapunov function candidate $V_{\theta}$. For notational convenience, we write $u$ to denote both the control function and the parameters that define the function. The learning process updates both the $\theta$ and $u$ parameters to improve the likelihood of satisfying the Lyapunov conditions, which we formulate as a cost function named the {\em Lyapunov risk}. 
%Define the output of $\ell^{th}$ hidden layer is a linear map from $n_{\ell-1}$ neurons to $n_{\ell}$ neurons, added with a bias vector $B_{\ell}$ and transformed using an activation function $\sigma_{\ell}$. Then, the output of $\ell^{th}$ layer can be written as the form $y_{\ell}=\sigma_{\ell}\left(W_{\ell} y_{\ell-1}+B_{\ell}\right)$, where $1 < \ell \leq M$, $W_{\ell}$ is a $n_{\ell} \times n_{\ell-1}$ weight matrix and $B_{\ell}$ is a $n_{\ell} \times 1$ bias vector. In learning procedure, a candidate neural CLF is constructed by above network architecture with one hidden layer and fixed activation function $tanh(\cdot)$, so the template for a candidate neural CLF takes the form... where $x = \left[x_{1},x_{2},\ldots,x_{n}\right]^{\mathrm{T}} \in \mathbb{R}^{n}$ is a state and $\theta$ is composed of a set of weight matrices and bias vectors. 
The Lyapunov risk measures the degree of violation of the following Lyapunov conditions, as shown in Proposition (\ref{def:lya}). First, the value of $V_{\theta}\left(x\right)$ is positive; Second, the value of the Lie derivative $\lie V_{\theta}\left(x\right)$ is negative; Third, the value of $V_{\theta}(0)$ is zero. Conceptually, the overall Lyapunov control design problem is about minimizing the minimax cost of the form
\[\inf_{\theta,u}\sup_{x\in \mathcal{D}}\bigg(\max(0,-V_{\theta}(x))+\max(0,\lie V_{\theta}(x))+V_{\theta}^{2}(0)\bigg).\]The difficulty in control design problems is that the violation of the Lyapunov conditions can not just be estimated, but needs to be fully guaranteed over all states in $\mathcal{D}$. Thus, we need to rely on global search with complete guarantee for the inner maximization part, which we delegate to the falsifier explained in Section 3.2. For the learning step, we define the following Lyapunov risk function. 
\begin{definition}[Lyapunov Risk]\label{def:risk}
Consider a candidate Lyapunov function $V_{\theta}$ for a controlled dynamical system $f_u$ from Definition~\ref{def:dyn}. The Lyapunov risk is a defined by the following function
{\begin{equation}
    L_{\rho}\left(\theta,u\right)= \mathbb{E}_{x\sim \rho(\mathcal{D})}\bigg(\max(0,-V_{\theta}(x))+\max(0,\lie V_{\theta}(x))+V_{\theta}^{2}(0)\bigg),
\end{equation}}where $x$ is a random variable over the state space of the system with a distribution $\rho$. %$\nabla_{f_{u}}V_{\theta}$ is the Lie derivative in Definition~\ref{def:lie}. 
In practice, we work with the Monte Carlo estimate named the {\em empirical Lyapunov risk} by drawing samples: 
{\begin{equation}
    L_{N,\rho}\left(\theta,u\right)= \frac{1}{N}\sum^{N}_{i=1}\bigg(\max(-V_{\theta}(x_{i}),0)+\max(0,\lie V_{\theta}(x_{i}))\bigg)+V_{\theta}^{2}(0),
\end{equation}}where $x_{i},1\leq i \leq N$ are samples of the state vectors sampled according to $\rho(\mathcal{D})$. %$V_{\theta}\left(x_{k}\right)$ is the output of network given $x_{k}$ and $\theta$ and $\nabla_{f}V_{\theta}\left(x_{k}\right)=\sum_{i=1}^{n} \frac{\partial V_{\theta}}{\partial x_{k_{i}}} f_{i}\left(x_{k},u_{q}\left(x_{k}\right)\right)$.
\end{definition}
It is clear that the empirical Lyapunov risk is an unbiased estimator of the Lyapunov risk function. It is clear that $L_{N,\rho}$ is an unbiased estimator of $L_{\rho}$. 

\begin{figure}[th!]    
    \centering
    \includegraphics[width=1\textwidth]{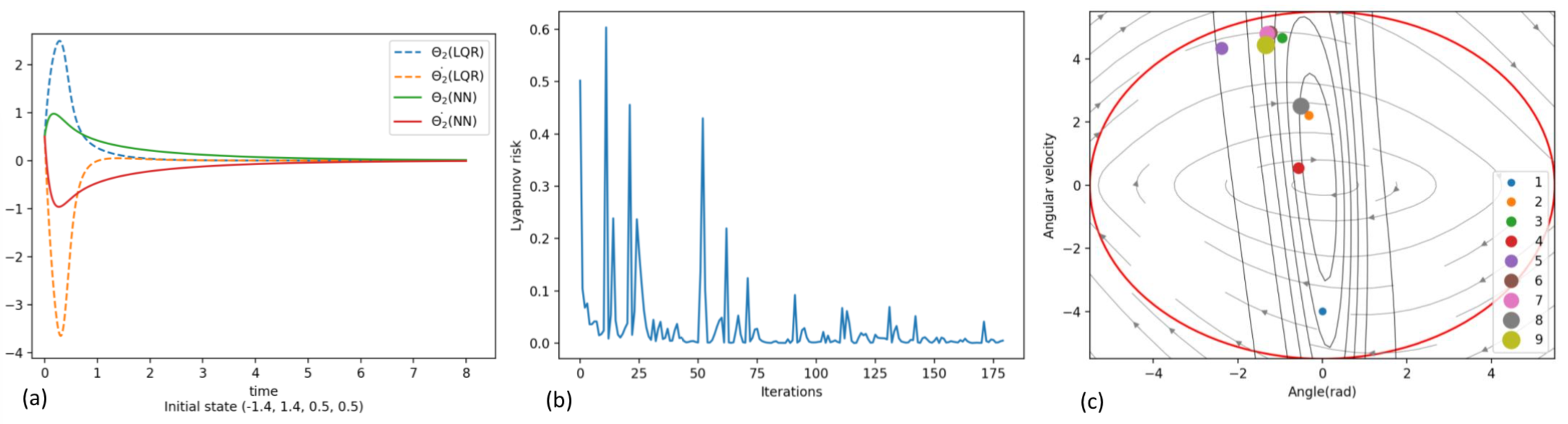}
    \caption{(a) Comparison between LQR and deep-learned controllers for 2-link planar robot balancing. (b) The Lyapunov risk decreases quickly over iterations. (c)  Counterexamples returned by falsifiers from several epochs, which quickly guides the learner to focus on sepcial regions in the space.}
    \label{fig:main1}
\end{figure}

Note that $L_{\rho}$ is positive semidefinite, and any $(\theta,u)$ that corresponds to a true Lyapunov function satisfies $L(\theta,u)$=0. Thus, Lyapunov functions define global minimizers of the Lyapunov risk function.
\begin{proposition}
Let $V_{\theta_o}$ be a Lyapunov function for dynamical system $f_{u_o}$ where $u_o$ is the control parameters. Then $(\theta_o,u_o)$ is a global minimizer for $L_{\rho}$ and $L_{\rho}(\theta_o,u_o)=0$. 
\end{proposition}
Note that both $V_{\theta}$ and $f_u$ are highly nonlinear (even though $u$ is almost always linear in practice), and thus $L(\theta,u)$ generates a highly complex landscape. Surprisingly, multilayer feedforward $\tanh$ networks and stochastic gradient descent can quickly produce generalizable Lyapunov functions with nice geometric properties, as we report in detail in the experiments. In Figure~\ref{fig:main1} (b), we show an example of how the Lyapunov risk is minimized over iterations on the inverted pendulum example. 
\paragraph{Initialization and improvement of control performance over LQR.} Because of the local nature of stochastic gradient descent, it is hard to learn good control functions through random initialization of control parameters. Instead, the parameters $u$ in the control function are initialized to the LQR solution, obtained for the linearized dynamics in a small neighborhood around the stationary point. On the other hand, the initialization of the neural network Lyapunov functions can be completely random. We observe that the final learned controller often delivers significantly better control solutions than the initalization from LQR. Figure~\ref{fig:main1}(a) shows how the learned control reduces oscillation of the system behavior in the 2-link planar robot balancing example and achieve more stable control.  

\subsection{Falsification and Counterexample Generation}
% Derivatives
For each control and Lyapunov function pair $(V_{\theta}, u)$ that the learner obtains, the falsifier's task is to find states that violate the Lyapunov conditions in Proposition~\ref{def:lya}. We formulate the {\em negations} of the Lyapunov conditions as a nonlinear constraint solving problem over real numbers. These {\em falsification constraints} are defined as follows.  
\begin{definition}[Lyapunov Falsification Constraints] Let $V$ be a candidate Lyapunov function for a dynamical system defined by $f_u$ defined in state space $\mathcal{D}$. Let $\varepsilon\in\mathbb{Q}^+$ be a small constant parameter that bounds the tolerable numerical error. The Lyapunov falsification constraint is the following first-order logic formula over real numbers:
{\small\[\Phi_{\varepsilon}(x):= \bigg(\sum_{i=1}^n x_i^2 \geq \varepsilon\bigg) \wedge \bigg(V(x)\leq 0 \vee \lie V(x)\geq 0\bigg)\]}where $x$ is bounded in the state space $\mathcal{D}$ of the system. The numerical error parameter $\varepsilon$ is explicitly introduced for controlling numerical sensitivity near the origin. Here $\varepsilon$ is orders of magnitude smaller than the range of the state variables, i.e., $\sqrt{\varepsilon}\ll\min(1,||\mathcal{D}||_{2})$. 
\label{def:constrt}
\end{definition}
\begin{remark}
The numerical error parameter $\varepsilon$ allows us to avoid pathological problems in numerical algorithms such as arithmetic underflow. Values inside this tiny ball correspond to disturbances that are physically insignificant. This parameter is important for eliminating from our framework the numerical sensitivity issues commonly observed in SOS/SDP methods. Also note the $\varepsilon$-ball does not affect properties of the Lyapunov level sets and regions of attraction outside it (i.e., $\mathcal{D}\setminus B_{\varepsilon}$).
\end{remark}

The falsifier computes solutions of the falsification constraint $\Phi_{\varepsilon}(x)$. Solving the constraints requires global minimization of a highly nonconvex functions (involving Lie derivatives of the neural network Lyapunov function), and it is a computationally expensive task (NP-hard). We rely on recent progress in nonlinear constraint solving in SMT solvers such as dReal~\cite{DBLP:conf/cade/GaoAC12}, which has been used for similar control design problems~\cite{Kapinski2014} that do not involve neural networks.  
\begin{example}
Consider a candidate Lyapunov function $V(x)=\tanh(a_1x_1+a_2x_2+b)$ and dynamics $\dot{x_1}=-x_2^2$ and $\dot{x_2}=\sin(x_1)$. The falsification constraint is of the following form
\begin{multline*}\Phi_{\varepsilon}(x):= (x_1^2+x_2^2)\geq \varepsilon \wedge (\tanh(a_1x_1+a_2x_2+b)\leq 0 \vee a_1(1-\tanh^2(a_1x_1+a_2x_2+b))(-x_2^2)
\\+a_2(1-\tanh^2(a_1x_1+a_2x_2+b))\sin(x_1)\geq 0))
\end{multline*}
which is a nonlinear non-polynomial disjunctive constraint system. The actual examples used in our experiments all use larger two-layer $\tanh$ networks and much more complex dynamics. 
\end{example}
To completely certify the Lyapunov conditions, the constraint solving step for $\Phi_{\varepsilon}(x)$ can never fail to report solutions if there is any. This requirement is rigorously proved for algorithms in SMT solvers such as dReal~\cite{DBLP:conf/cade/GaoKC13}, as a property called delta-completeness~\cite{DBLP:conf/cade/GaoAC12}. 
\begin{definition}[Delta-Complete Algorithms]
Let $C$ be a class of quantifier-free first-order constraints. Let $\delta\in\mathbb{Q}^+$ be a fixed constant. We say an algorithm $\mathcal A$ is $\delta$-complete for $C$, if for any $\varphi(x)\in C$, $\mathcal A$ always returns one of the following answers correctly: $\varphi$ does not have a solution (unsatisfiable), or there is a solution $x=a$ that satisfies $\varphi^{\delta}(a)$. Here, $\varphi^{\delta}$ is defined as a small syntactic variation of the original constraint (precise definitions are in~\cite{DBLP:conf/cade/GaoAC12}). 
\end{definition}
In other words, if a delta-complete algorithm concludes that a formula $\Phi_{\varepsilon}(x)$ is unsatisfiable, then it is guaranteed to not have any solution. In our context, this is exactly what we need for ensuring that the Lyapunov condition holds over all state vectors. If $\Phi_{\varepsilon}(x)$ is determined to be $\delta$-satisfiable, we obtain counterexamples that are added to the training set for the learner. Note that the counterexamples are simply state vectors without labels, and their Lyapunov risk will be determined by the learner, not the falsifier. Thus, although it is possible to have spurious counterexamples due to the $\delta$ error, they are used as extra samples and do not harm correctness of the end result. In all, when delta-complete algorithms in dReal return that the falsification constraints are unsatisfiable, we conclude that the Lyapunov conditions are satisfied by the candidate Lyapunov and control functions. Figure ~\ref{fig:main1}(c) shows a sequence of  counterexamples found by the falsifier to improve the learned results. 
\begin{remark}
When solving $\Phi_{\varepsilon}(x)$ with $\delta$-complete constraint solving algorithms, we use $\delta\ll\varepsilon$ to reduce the number of spurious counterexamples. Following delta-completeness, the choice of $\delta$ does not affect the guarantee for the validation of the Lyapunov conditions. 
\end{remark}
%To summarize, in the falsification step, we either conclude that the Lyapunov and control functions $\langle V_{\theta},u\rangle$ fully satisfy the Lyapunov conditions (when the falsification constraints are unsatisfiable), or obtain counterexamples that show violations of the Lyapunov conditions. In the latter case, we can optionally augment the counterexample by sampling a small neighborhood around it to increase its statistical significance in the training set. The falsifier returns these new samples to the learner for the next iteration of Lyapunov risk minimization. 

\subsection{Tuning Region of Attraction}

\begin{figure}[th!]    
    \centering
    \includegraphics[width=1\textwidth]{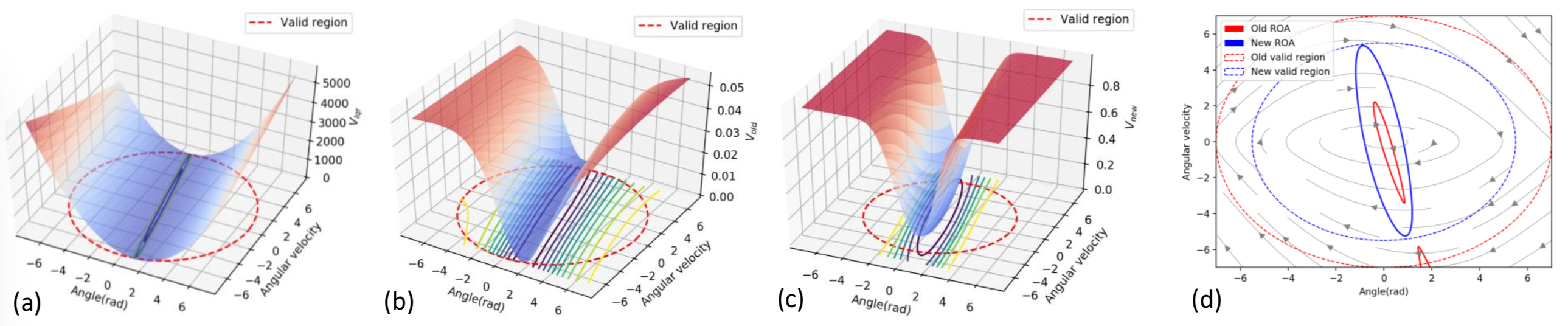}
    \caption{(a) Lyapunov function found by the initial LQR controller. (b) Lyapunov function found by learning without tuning the ROA. (c) Lyapunov function found by learning after adding the ROA tuning term. (d) Comparison of ROA for the different Lyapunov functions.}
    \label{fig:inverted}
\end{figure}

%\todo[inline]{explain how to add additional cost terms to manipulate the shape of the Lyapunov level sets. Emphasize that this tuning is only achievable by our methods, whereas no tuning like this can be done in SOS methods. }
An important feature of the proposed learning framework is that we can adjust the cost functions to learn control and Lyapunov functions favoring various additional properties. In fact, the most practically important performance metric for a stabilizing controller is its region of attraction (ROA). An ROA defines a forward invariant set that is guaranteed to contain all possible trajectories of the system, and thus can conclusively prove safety properties. Note that the Lyapunov conditions themselves do not directly ensure safety, because the system can deviate arbitrarily far before coming back to the stable equilibrium. Formally, the ROA of an asymptotically stable system is defined as: 
%Now we can prove the stability of the stable equilibrium of dynamic within the valid region verified by dReal. Any initial state start inside the valid region can be assure that its trajectory will approach the stable equilibrium and stay in the $\varepsilon$-neighborhood of the equilibrium. However, ensuring whole trajectories contained in certain region is also important for real applications. Since the conditions in $(2)$ are not sufficient to prove the whole trajectory will stay in the valid region all the time, we consider the task of estimating the region of attraction (ROA).
\begin{definition}[Region of Attraction] Let $f_u$ define a system asymptotically stable at the origin with Lyapunov function $V$ for domain $\mathcal{D}$. A region of attraction $R$ is a subset of $\mathcal{D}$ that contains the origin and guarantees that the system never leaves $R$. Any level set of $V$ completely contained in $\mathcal{D}$ defines an ROA. That is, 
%The region of attraction of the stationary point $\bar{x}$ is a subset of the state space defined as $\mathcal{R}_{A}=\{ x \in \mathcal{R}^{n}:\lim _{t \rightarrow \infty}|x(t)-\bar{x}(t)|=0\}$. Assume there exist a Lyapunov function $V$ such that the condition in $(2)$ hold on region $\mathcal{D}$ containing the stationary point $\bar{x}$, and 
for $\beta>0$, if $R_{\beta}=\{V(x)\leq \beta\}\subseteq \mathcal{D}$, then $R_{\beta}$ is an ROA for the system.
\end{definition}
To maximize the ROA produced by a pair of Lyapunov function and control function, we add a cost term to the Lyapunov risk that regulates how quickly the Lyapunov function value increases with respect to the radius of the level sets, by using $L_{N,p}(\theta,u)+\frac{1}{N}\sum^{N}_{i=1}\|x_{i}\|_{2}-\alpha V_{\theta}(x_{i})$ following Definition~\ref{def:risk}. Here $\alpha$ is tunable parameter. We observe that the regulator can have major effect on the performance of the learned control functions. Figure \ref{fig:inverted} illustrates such an example, showing how different control functions are obtained by regulating the Lyapunov risk to achieve larger ROA.
% We estimate the region of attraction by using Lyapunov functions. Assume there exist a Lyapunov function $V$ such that the condition in $(2)$ hold on region $\mathcal{D}$, and if $\mathcal{D}_{\beta} =\{ x \in \mathcal{D}:V(x)\leq \beta\}$ is bounded, then $\mathcal{D}_{\beta}$ is a region of attraction. By the Definition $8$, as long as the initial states start inside $\mathcal{D}_{\beta}$, their whole trajectories will never leave $\mathcal{D}_{\beta}$.
% It is note that the level cure of Lyapunov functions yield the restriction of the shape of ROA. However, our method is flexible to shape the Lyapunov level cure while constructing. 

\begin{algorithm}[h!b]
    \caption{Neural Lyapunov Control}\label{alg:clf}
    \begin{algorithmic}[1]
    \Function{Learning}{$X,f,q^{lqr}$}
        % \State Set learning rate $(0.01)$, input dimension $(D_{in}=\#\text{ of state variables})$, hidden dimension  \indent $(D_{h})$ and output dimension $(D_{out}=1)$
        \State Set learning rate $(0.01)$, input dimension $(\#\text{ of state variables})$, output dimension $(1)$
        % \State Initialize neural Lyapunov function $(V)$ with random weights $\theta$
        \State Initialize feedback controller $u$ to LQR solution $q^{lqr}$
        \State \textbf{Repeat:}
        \State \indent  $V_{\theta}\left(x\right),u\left(x\right) \leftarrow \text{NN}_{\theta,u}\left(x\right)$ \Comment{Forward pass of neural network}
        \State \indent$\lie V_{\theta}\left(x\right)\leftarrow \sum_{i=1}^{D_{in}} \frac{\partial V}{\partial x_{i}} [f_{u}]_i(x)$
        \State \indent Compute Lyapunov risk $L\left(\theta,u\right)$
        \State \indent $\theta\leftarrow \theta+\alpha \nabla_{{\theta}} L\left(\theta,u\right)$
        \State \indent $u\leftarrow u+\alpha \nabla_{{u}} L\left(\theta,u\right)$  \Comment{Update weights using SGD}
        \State \textbf{Until} convergence
        \State \Return  $V_{\theta},u$
    \EndFunction
        \Function{Falsification}{$f,u,V_{\theta},\varepsilon,\delta$}
         \State Encode conditions in Definition \ref{def:constrt} 
         \State Using SMT solver with $\delta$ to verify the conditions
         \State \Return satisfiability
    \EndFunction
     \Function{Main}{ }
     
         \State \textbf{Input:} dynamical system $(f)$, parameters of LQR $(q^{lqr})$, radius $(\varepsilon)$, precision $(\delta)$ and an \indent initial set of randomly sampled states in $D$
         \While{Satisfiable}
             \State Add counterexamples to $X$
             \State $V_{\theta},u\leftarrow$ \Call{Learning}{$X,f,q^{lqr}$}
             \State CE$\leftarrow$ \Call{Falsification}{$f,u,V_{\theta},\varepsilon,\delta$}
         \EndWhile
     \EndFunction
\end{algorithmic}
\end{algorithm}
\section{Experiments}
%\todo[inline]{Experiements: Set dReal region to be ellipsoid during tuning, based on shape observed from previously found level sets}
%\todo[inline]{need the following plots for inverted pendulum. 1. compare with ROA of SOS. 2. the difference in the convergence behavior of points in and outside the ROA. (mention that it is similar to what has been found in the Lyapunov network paper) 3. I think we can p1ot the Lie derivative too}
We demonstrate that the proposed methods find provably stable control and Lyapunov functions on various nonlinear robot control problems. In all the examples, we use a learning rate of $0.01$ for the learner, an $\varepsilon$ value of $0.25$ and $\delta$ value of $0.01$ for the falsifier, and re-verify the result with smaller $\varepsilon$ in Table~\ref{tab:roa}. We emphasize that the choices of these parameters do not affect the stability guarantees on the final design of the control and Lyapunov functions. We show that the region of attraction is enlarged by 300\% to 600\% compared to LQR results in these examples. Full details of the results and system dynamics are provided in the Appendix. Note that for the Caltech ducted fan and n-link balancing examples, we numerically relaxed the conditions slightly when the learning has converged, so that the SMT solver dReal does not run into numerical issues. More details on the effect of such relaxation can be found in the paper website~\cite{website}. 

\begin{table}[th]
\centering
\begin{tabular} {|c||c|c|c|c|c|c|c|c|}
\hline
Benchmarks & Learning time & falsification time  & \# samples & \# iterations & $\varepsilon$ \\
\hline
{Inverted Pendulum} & 25.5 & 0.6 & 500 & 430 & 0.04 \\
% \hline
% {Cart-Pole}& 384 & 5.08 & 1800 & 1500 \\
\hline
{Path Following} & 36.3 & 0.2 & 500 & 610 & 0.01 \\
\hline
{Caltech Ducted Fan} & 1455.16 & 50.84 & 1000 & 3000 & 0.01\\
\hline
{2-Link Balancing} & 6000 & 458.27 & 1000 & 4000 & 0.01\\
\hline
\end{tabular}
\vspace{0.3cm}
\caption{Runtime statistics of the full procedures on four nonlinear control examples.}
\label{tab:roa}
\end{table}

\begin{figure}[th!]
    \centering
    \includegraphics[width=1\textwidth]{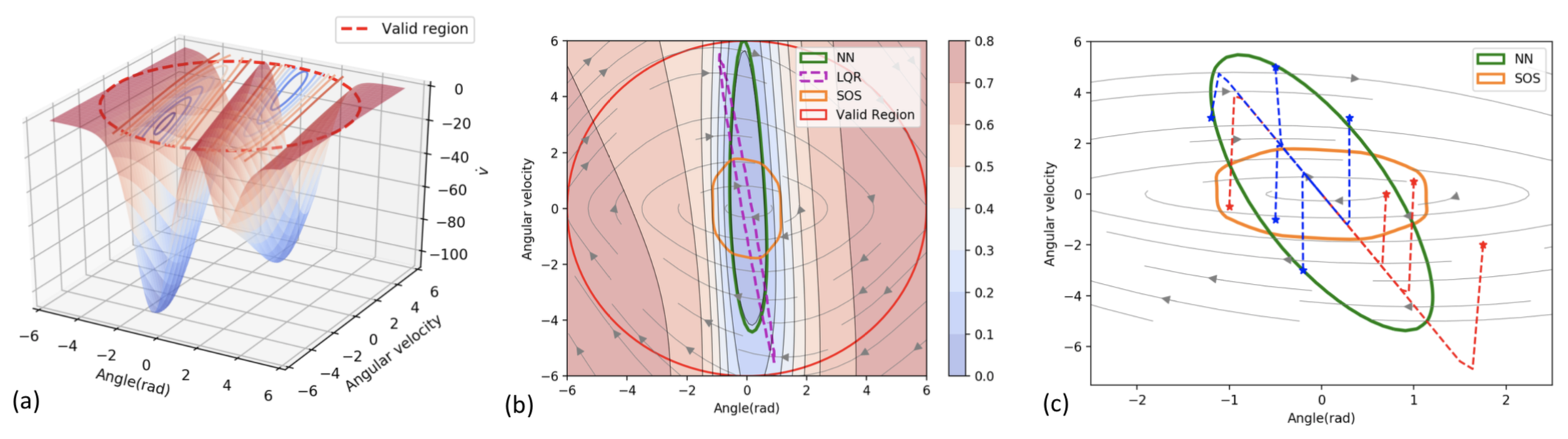}
    % \caption{Show our Lyapunov function, and Vdot, and then comparison of ROAs.}
    \caption{Results of Lyapunov functions for inverted pendulum. (a) Lie derivative of learned Lyapunov function over valid region. Its value is negative over the valid region, satisfying the Lyapunov conditions. (b) ROA estimated by different Lyapunov functions. Our method enlarges the ROA from LQR three times. (c) Validation of ROAs. Stars represent initial states. It shows trajectories start near border of the ROA defined by the learned neural Lyapunov function are safely bounded within the green region. On the contrary, many trajectories (red) starting inside the SOS region can escape, and thus the region fails to satisfy the ROA properties.}
    \label{fig:roa}
\end{figure}

%Several nonlinear dynamical systems will be presented to show our approach can efficiently construct Lyapunov functions and obtain large region of attraction.
\paragraph{Inverted pendulum.} The inverted pendulum is a standard nonlinear control problem for testing different control methods. This system has two state variables, the angular position $\theta$, angular velocity $\dot{\theta}$ and one control input $u$.
%$\theta$ and $\dot{\theta}$ represent the angular position from the inverted position and angular velocity with dynamics:
%\begin{equation}
%\ddot{\theta} = m^{-1}\ell^{-2}(mg\ell\sin{\left(\theta\right)}+u-0.1\dot{\theta})
%\end{equation}
%Using constants $g=9.81$, $m=0.15$ and $\ell=0.5$, 
Our learning procedure finds a neural Lyapunov function that is proved to be valid within the domain $\|x\|_{2} \leq 6$. In contrast, the ROA found by SOS/SDP techniques is an ellipse with large diameter of $1.75$ and short diameter of $1.2$. Using LQR control on the linearized dynamics, we obtain an ellipse with large diameter of $6$ and short diameter of $0.1$. We observe that among all the examples in our experiments, this is the only one where the SOS Lyapunov function has passed the complete check by the constraint solver, so that we can compare to it. The Lyapunov function obtained by LQR gives a larger ROA if we ignore the linearization error. The different regions of attractions are shown in Figure~\ref{fig:roa}. These values are consistent with the approximate maximum region of attraction reported in~\cite{Spencer2018}. In particular, Figure~\ref{fig:roa} (c) shows that the SOS function does not define a big enough ROA, as many trajectories escape its region. 

\begin{figure}[th!]
    \centering
    \includegraphics[width=1\textwidth]{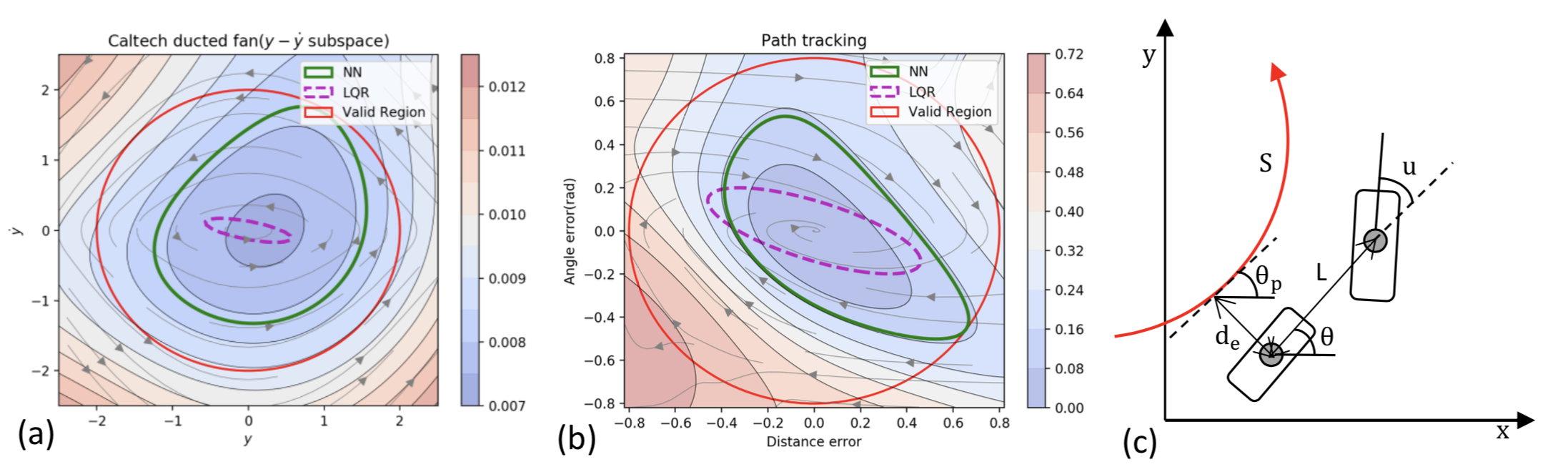}
    \caption{(a) Comparison of ROAs for Caltech ducted fan. (b) Comparison of ROAs for path following. (c) Schematic diagram of wheeled vehicle to show the nonlinear dynamics.}
    \label{fig:fig4}
\end{figure}

\textbf{Caltech ducted fan in hover mode.} The system describes the motion of a landing aircraft in hover mode with two forces $u_{1}$ and $u_{2}$. The state variables $x$, $y$, $\theta$ denote the position and orientation of the centre of the fan. There are six state variables $[x , y, \theta, \dot{x}, \dot{y}, \dot{\theta}]$. %We find the control function
%\[\begin{aligned} 
%u_{1} &= 0.5000x +0.000002y -2.1339\theta +2.7899\dot{x} -0.00000003\dot{y}-1.3992\dot{\theta}\\
%u_{2} &= 0.000001x -1.0000y -0.000003\theta -0.000003\dot{x} -5.0407\dot{y}-0.000001\dot{\theta}.
%\end{aligned}\]
The dynamics, neural Lyapunov function with two layers of $\tanh$ activation functions, and the control policy are given in the Appendix. 
%The function is guaranteed to satisfy Lyapunov conditions in the region $\left\|x\right\|_{L_2}\leq 2$ under precision $\delta=0.01$. 
In Figure~\ref{fig:fig4}$(a)$, we show that the ROA is significantly larger than what can be obtained from LQR. 

\paragraph{Wheeled vehicle path following.} We consider the path tracking control using kinematic bicycle model (see Figure~\ref{fig:fig4}$(c)$). We take the angle error $\theta_{e}$ and the distance error $d_{e}$ as state variables. Assume a target path is a unit circle, then we obtain the Lyapunov function within $\left\|x\right\|_{2} \leq 0.8$.
%  $V=\tanh(W_2\tanh(W_{1}x+B_{1})+B_{2})$, where $x=[d_{e}\ \theta_{e}]^{T}$ and 
% {\small\[W_{1}=
% \begin{bmatrix*}[r]
% -0.0059 & 0.0597 & -0.0053 & -0.0249 & 0.0021\\
% -0.0053 & -0.0008 & 0.0100 & -0.0021 & 0.0001\\
% \end{bmatrix*}^T,\]
% \[W_{2}= 
% \begin{bmatrix}
% 0.5371 & 0.1276 & 0.2407 & 0.1228 & 0.3277\\ 
% \end{bmatrix},\]
% \[B_{1}=
% \begin{bmatrix}
% -0.9597& -0.8179 & -0.6969 &-0.6101 & 1.1636\\
% \end{bmatrix} \text{ and } 
% B_{2}=
% \begin{bmatrix}0.9767\\ 
% \end{bmatrix}\]}
% and the feedback controller is $u = -0.8471d_{e} -1.6414\theta_{e}$.

% \begin{figure}[t!]
%     \centering
%     \includegraphics[width=1\textwidth]{Fig5.png}
%     \caption{Results of humanoid balance. (a) Schematic diagram. (b) Learned Lyapunov function. (c) Lie derivative of Lyapunov function. (d) Comparison of the region of attraction. }
%     \label{fig:2link}
% \end{figure}

\paragraph{N-Link Planar Robot Balancing.} The $n$-link pendulum system has $n$ control inputs and $2n$ state variables $[\theta_{1},\theta_{2},\ldots,\theta_{n},\dot{\theta_{1}},\dot{\theta_{2}},\ldots,\dot{\theta_{n}}]$, representing the $n$ link angles and $n$ angle velocities. Each link has mass $m_{i}$ and length $\ell_{i}$, and the moments of inertia $I_{i}$ are computed from the link pivots, where $i=1,2, \ldots,n$.  
% The dynamics has the following form: \begin{equation}
% M\left(\theta\right) \ddot{\theta}+C(\theta, \dot{\theta}) \dot{\theta}+ \tau(\theta)=Bu,
% \end{equation}
% where
% {\small\[
% \begin{aligned}
% \theta &= \left[\theta_{1},\theta_{2},\ldots,\theta_{n}\right]^{\mathrm{T}} \in \mathbb{R}^{n}, M(\theta)=\left[a_{i j} \cos \left(\theta_{j}-\theta_{i}\right)\right], M\left(\theta\right) \in \mathbb{R}^{n\times n}\\
% C(\theta, \dot{\theta})&=\left[-a_{i j} \dot{\theta}_{j} \sin \left(\theta_{j}-\theta_{i}\right)\right], C(\theta, \dot{\theta})\in \mathbb{R}^{n\times n}, \tau(\theta)=\left[-b_{i} \sin \theta_{i}\right],G(\theta)\in \mathbb{R}^{n},\\
% %B &= \left[1,1,\ldots,1\right]^{\mathrm{T}}
% \end{aligned}
% \]
% \[
% \left\{\begin{array}{l}{a_{i i}=I_{i}+m_{i} \ell_{c i}^{2}+\ell_{i}^{2} \sum_{k=i+1}^{n} m_{k}, 1 \leq i \leq n} \\ {a_{i j}=a_{j i}=m_{j} \ell_{i} \ell_{c j}+\ell_{i} \ell_{j} \sum_{k=j+1}^{n} m_{k}, 1 \leq i<j \leq n}\end{array}\right.
% b_{i}=\left(m_{i} \ell_{c i}+\ell_{i} \sum_{k=i+1}^{n} m_{k}\right) g, 1 \leq i \leq n,\]}
We find a neural Lyapunov function for the 2-link pendulum system within $\left\|x\right\|_{2} \leq 0.5$.
% The Lyapunov function is
% $V=\tanh(W_2\tanh(W_{1}x+B_{1})+B_{2})$, where $x=[\theta_{1}\ \theta_{2}\ \theta_{3}\ \dot{\theta_{1}}\ \dot{\theta_{2}}\ \dot{\theta_{3}}]^{T}$ and
% {\small
% \[W_{1}=
% \begin{bmatrix*}[r]
% -0.1919 & 0.1715& -0.0481 & 0.0707 & 0.1923 & 0.0548\\
% 0.0943 & 0.0112 & 0.0027 & 0.0102 & -0.0005 & 0.0002\\
% 0.0942 & -0.2393 & 0.0932 & -0.0692 & -0.1582 & 0.0002\\
% 0.0943 & 0.0112 & 0.0027 & 0.0102 & -0.0005 & -0.0221\\
% -0.1136& -0.1927 & -0.0753 & -0.0407 & -0.1289 & 0.0246\\
% -0.1645& 0.2017 & 0.0412 & -0.1091 & -0.1892 &  -0.1396\\
% 0.0868& 0.0103 & 0.0030 &  0.0094 & -0.0002 & -0.0007\\
% \end{bmatrix*}^T,\]
% \[W_{2}= 
% \begin{bmatrix}
% 0.0017 & 0.4299 & 0.0023 &  -0.0021 & 0.0002 & -0.5047\\ 
% \end{bmatrix},\]
% \[B_{1}=
% \begin{bmatrix}
% -0.5246& -0.3993 & -0.3698 &0.1214 & 0.2343 & 0.4633\\
% \end{bmatrix} \text{ and } 
% B_{2}=
% \begin{bmatrix}0.3918\\ 
% \end{bmatrix},\]}
% and the control functions are
% \[\begin{aligned} 
% u_{1} &= -101.7856\theta_{1} -8.9265\theta_{2} -3.467\theta_{3}-28.5081\dot{\theta_{1}} -14.0951\dot{\theta_{2}}-7.3643\dot{\theta_{3}}\\
% u_{2} &= 15.8736\theta_{1}-62.5769\theta_{2} -4.0104\theta_{3}-7.8591\dot{\theta_{1}}  -12.6341\dot{\theta_{2}}-7.3690\dot{\theta_{3}}\\
% u_{3} &= 5.1672\theta_{1} +7.2750\theta_{2} -42.4820\theta_{3}-2.6997\dot{\theta_{1}} -4.9186\dot{\theta_{2}}-11.8446\dot{\theta_{3}}
% \end{aligned}\]
In Figure 5, we show the shape of the neural Lyapunov functions on two of the dimensions, and the ROA that the control design achieves. We also provide a video of the control on the 3-link model. 
\begin{figure}[h!]
    \centering
    \includegraphics[width=1\textwidth]{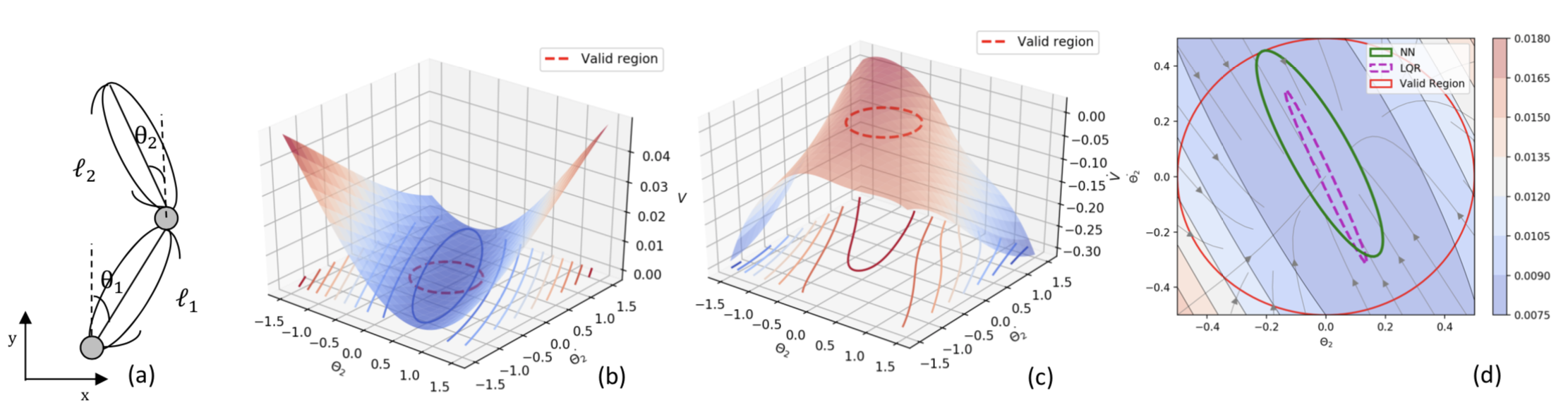}
    \caption{Results of n-link planar robot balancing. (a) Schematic diagram. (b) Learned Lyapunov function. (c) Lie derivative of Lyapunov function. (d) Comparison of the region of attraction. }
    \label{fig:2link}
\end{figure}
\section{Conclusion}
We proposed new methods to learn control policies and neural network Lyapunov functions for highly nonlinear systems with provable guarantee of stability. The approach significantly simplifies the process of nonlinear control design, provides end-to-end provable correctness guarantee, and can obtain much larger regions of attraction compared to existing control methods. We show experiments on challenging nonlinear problems central to various nonlinear control problems. The proposed methods demonstrate clear advantage over existing methods. We envision that neural networks and deep learning will lead to better solutions to core problems in robot control design. 

\newpage

\bibliographystyle{plain}
\bibliography{relu}

\begin{thebibliography}{10}

\bibitem{Ahmadi2015}
Amir~A. Ahmadi and Rapha{\"{e}}l~M. Jungers.
\newblock Lower bounds on complexity of lyapunov functions for switched linear
  systems.
\newblock {\em CoRR}, abs/1504.03761, 2015.

\bibitem{Ahmadi2011}
Amir~A. {Ahmadi}, M.~{Krstic}, and P.~A. {Parrilo}.
\newblock {}a globally asymptotically stable polynomial vector field with no
  polynomial lyapunov function.
\newblock In {\em 2011 50th IEEE Conference on Decision and Control and
  European Control Conference}, 2011.

\bibitem{Ahmadi2013}
Amir~A. Ahmadi and Pablo~A. Parrilo.
\newblock Stability of polynomial differential equations: Complexity and
  converse lyapunov questions.
\newblock {\em CoRR}, abs/1308.6833, 2013.

\bibitem{Ahmadi2012}
Amir~Ali Ahmadi.
\newblock On the difficulty of deciding asymptotic stability of cubic
  homogeneous vector fields.
\newblock In {\em American Control Conference, {ACC} 2012, Montreal, QC,
  Canada, June 27-29, 2012}, pages 3334--3339, 2012.

\bibitem{DBLP:journals/corr/AmodeiOSCSM16}
Dario Amodei, Chris Olah, Jacob Steinhardt, Paul~F. Christiano, John Schulman,
  and Dan Man{\'{e}}.
\newblock Concrete problems in {AI} safety.
\newblock {\em CoRR}, abs/1606.06565, 2016.

\bibitem{Berkenkamp2016}
F.~{Berkenkamp}, R.~{Moriconi}, A.~P. {Schoellig}, and A.~{Krause}.
\newblock Safe learning of regions of attraction for uncertain, nonlinear
  systems with gaussian processes.
\newblock In {\em 2016 IEEE 55th Conference on Decision and Control (CDC)},
  pages 4661--4666, Dec 2016.

\bibitem{Berkenkamp2017}
Felix Berkenkamp, Matteo Turchetta, Angela Schoellig, and Andreas Krause.
\newblock Safe model-based reinforcement learning with stability guarantees.
\newblock In I.~Guyon, U.~V. Luxburg, S.~Bengio, H.~Wallach, R.~Fergus,
  S.~Vishwanathan, and R.~Garnett, editors, {\em Advances in Neural Information
  Processing Systems 30}, pages 908--918. Curran Associates, Inc., 2017.

\bibitem{website}
Ya-Chien Chang, Nima Roohi, and Sicun Gao.
\newblock Neural {L}yapunov control (project website),
  \url{https://yachienchang.github.io/NeurIPS2019}, 2019.

\bibitem{Chesi2009}
G.~{Chesi} and D.~{Henrion}.
\newblock Guest editorial: Special issue on positive polynomials in control.
\newblock {\em IEEE Transactions on Automatic Control}, 54(5):935--936, May
  2009.

\bibitem{Chow2018}
Yinlam Chow, Ofir Nachum, Edgar Duenez-Guzman, and Mohammad Ghavamzadeh.
\newblock A lyapunov-based approach to safe reinforcement learning.
\newblock In S.~Bengio, H.~Wallach, H.~Larochelle, K.~Grauman, N.~Cesa-Bianchi,
  and R.~Garnett, editors, {\em Advances in Neural Information Processing
  Systems 31}, pages 8092--8101. Curran Associates, Inc., 2018.

\bibitem{DBLP:conf/cade/GaoAC12}
Sicun Gao, Jeremy Avigad, and Edmund~M. Clarke.
\newblock Delta-{C}omplete decision procedures for satisfiability over the
  reals.
\newblock In {\em Automated Reasoning - 6th International Joint Conference,
  {IJCAR} 2012, Manchester, UK, June 26-29, 2012. Proceedings}, pages 286--300,
  2012.

\bibitem{DBLP:conf/cade/GaoKC13}
Sicun Gao, Soonho Kong, and Edmund~M. Clarke.
\newblock d{R}eal: An {SMT} solver for nonlinear theories over the reals.
\newblock In {\em Automated Deduction - {CADE-24} - 24th International
  Conference on Automated Deduction, Lake Placid, NY, USA, June 9-14, 2013.
  Proceedings}, pages 208--214, 2013.

\bibitem{lyapunovbook}
Wassim Haddad and Vijaysekhar Chellaboina.
\newblock Nonlinear dynamical systems and control: A lyapunov-based approach.
\newblock {\em Nonlinear Dynamical Systems and Control: A Lyapunov-Based
  Approach}, 01 2008.

\bibitem{Henrion2005}
D.~Henrion and A.~Garulli.
\newblock {\em Positive Polynomials in Control}, volume 312 of {\em Lecture
  Notes in Control and Information Sciences}.
\newblock Springer Berlin Heidelberg, 2005.

\bibitem{Jarvis2003}
Z.~{Jarvis-Wloszek}, R.~{Feeley}, {Weehong Tan}, {Kunpeng Sun}, and
  A.~{Packard}.
\newblock Some controls applications of sum of squares programming.
\newblock In {\em 42nd IEEE International Conference on Decision and Control
  (IEEE Cat. No.03CH37475)}, volume~5, pages 4676--4681 Vol.5, Dec 2003.

\bibitem{Kapinski2014}
James Kapinski, Jyotirmoy~V. Deshmukh, Sriram Sankaranarayanan, and Nikos
  Arechiga.
\newblock Simulation-guided lyapunov analysis for hybrid dynamical systems.
\newblock In {\em Proceedings of the 17th International Conference on Hybrid
  Systems: Computation and Control}, HSCC '14, pages 133--142. ACM, 2014.

\bibitem{Kwakernaak}
Huibert Kwakernaak.
\newblock {\em Linear Optimal Control Systems}.
\newblock John Wiley \& Sons, Inc., New York, NY, USA, 1972.

\bibitem{nlinkrobot}
Yannian Liu and Xin Xin.
\newblock Controllability and observability of n-link planar robot with a
  single actuator having different actuator-sensor configurations.
\newblock {\em IEEE Transactions on Automatic Control}, 61:1--1, 12 2015.

\bibitem{Lofberg2009}
Johan L{\"{o}}fberg.
\newblock Pre- and post-processing sum-of-squares programs in practice.
\newblock {\em IEEE Transactions on Automatic Control}, 54(5):1007--1011, 2009.

\bibitem{Majumdar2017}
Anirudha Majumdar and Russ Tedrake.
\newblock Funnel libraries for real-time robust feedback motion planning.
\newblock {\em The International Journal of Robotics Research}, 36(8):947--982,
  2017.

\bibitem{simple-linear}
Horia Mania, Aurelia Guy, and Benjamin Recht.
\newblock Simple random search of static linear policies is competitive for
  reinforcement learning.
\newblock In S.~Bengio, H.~Wallach, H.~Larochelle, K.~Grauman, N.~Cesa-Bianchi,
  and R.~Garnett, editors, {\em Advances in Neural Information Processing
  Systems 31}, pages 1805--1814. 2018.

\bibitem{Parrilo2000}
Pablo~A. Parrilo.
\newblock {\em Structured semidefinite programs and semialgebraic geometry
  methods in robustness and optimization}.
\newblock PhD thesis, California Institute of Technology, 2000.

\bibitem{Rasmussen2006}
C.E. Rasmussen and C.K.I. Williams.
\newblock {\em Gaussian Processes for Machine Learning}.
\newblock Adaptative computation and machine learning series. University Press
  Group Limited, 2006.

\bibitem{Ravanbakhsh2019}
Hadi Ravanbakhsh and Sriram Sankaranarayanan.
\newblock Learning control lyapunov functions from counterexamples and
  demonstrations.
\newblock {\em Autonomous Robots}, 43(2):275--307, 2019.

\bibitem{Spencer2018}
Spencer~M. Richards, Felix Berkenkamp, and Andreas Krause.
\newblock The lyapunov neural network: Adaptive stability certification for
  safe learning of dynamical systems.
\newblock In {\em Proceedings of The 2nd Conference on Robot Learning},
  volume~87 of {\em Proceedings of Machine Learning Research}, pages 466--476,
  29--31 Oct 2018.

\bibitem{russbook}
Russ Tedrake.
\newblock {\em Underactuated Robotics: Algorithms for Walking, Running,
  Swimming, Flying, and Manipulation (Course Notes for MIT 6.832).}
\newblock 2019.

\end{thebibliography}

\end{document}